\documentclass{article}
\usepackage{spconf,amsmath,graphicx}

\usepackage[numbers,sort]{natbib}

\title{
A Neural Network with Local Learning Rules \\ for Minor Subspace Analysis}
\name{Author(s) Name(s)}
\address{Author Affiliation(s)}
\twoauthors
 {Yanis Bahroun $^{\dagger}$}
	{$^{\dagger}$Center for Computational Neuroscience \\
	Flatiron Institute, Simons Foundation \\
	New York, NY USA}
 {Dmitri B. Chklovskii $^{\dagger \star}$}
	{$^{\star}$Neuroscience Institute \\
	NYU Langone Medical Center \\
	New York, NY USA}

\usepackage{hyperref}       
\usepackage{url}            
\usepackage{booktabs}       
\usepackage{amsfonts}       
\usepackage{nicefrac}       
\usepackage{microtype}      
\usepackage{enumitem}
\usepackage{sidecap}

\usepackage{multicol}
\usepackage{amsmath}
\usepackage{graphicx}

\usepackage{algorithm}

\usepackage[noend]{algpseudocode}

\usepackage{amssymb}
\usepackage{graphicx}
\usepackage{amssymb,amsmath,array}
\usepackage{amsmath,amsfonts,amsthm,bm} 

\usepackage{tikz}
\usetikzlibrary{arrows}
\usepackage{epstopdf}
\usepackage{times}

\usepackage{graphicx}
\usepackage{pgfgantt}

\usepackage{subfig}
\usepackage[graphicx]{realboxes}

\usepackage{amsmath}
\usepackage{graphicx}
\usepackage{caption}
\usepackage{bbm}
\usepackage{datetime}
\usepackage[colorinlistoftodos]{todonotes}
\usepackage{color}
\definecolor{dblue}{rgb}{0,0,0.6}
\usepackage{amssymb}
\usepackage{relsize}
\usepackage{pifont}
\usepackage{multirow}

\usepackage{pifont}
\usepackage{multicol}
\usepackage{graphicx}
\usepackage{mathtools}
\usepackage{amssymb}
\usepackage{amsthm}
\usepackage{wrapfig,hyperref}
\hypersetup{colorlinks=true,linkcolor=blue,citecolor=red}

\usepackage{natbib}
\usepackage{longtable}

\newtheorem{prop}{Proposition}
\newtheorem*{prop*}{Proposition}

\newtheorem{propx}{Proposition}

\theoremstyle{definition}

\DeclareMathOperator{\tr}{Tr}
\DeclareMathOperator{\Tr}{Tr}

\newcommand{\norm}[1]{\lVert{#1}\rVert}

\renewcommand{\u}{{\bf u}}
\renewcommand{\v}{{\bf v}}
\newcommand{\w}{{\bf w}}
\newcommand{\x}{{\bf x}}
\newcommand{\y}{{\bf y}}

\newcommand{\C}{{\bf C}}

\newcommand{\F}{{\bf F}}

\newcommand{\I}{{\bf I}}

\newcommand{\M}{{\bf M}}

\newcommand{\R}{\mathbb{R}}

\newcommand{\U}{{\bf U}}
\newcommand{\V}{{\bf V}}
\newcommand{\W}{{\bf W}}
\newcommand{\X}{{\bf X}}
\newcommand{\Y}{{\bf Y}}

\renewcommand{\v}{{\bf v}}

\newcommand{\Real}{{\mathbb R}}

\begin{document}
%
\maketitle
\begin{abstract}
The development of neuromorphic hardware and modeling of biological neural networks requires algorithms with local learning rules. Artificial neural networks using local learning rules to perform principal subspace analysis (PSA) and clustering have recently been derived from principled objective functions. However, no biologically plausible networks exist for minor subspace analysis (MSA), a fundamental signal processing task. MSA extracts the lowest-variance subspace of the input signal covariance matrix. Here, we introduce a novel similarity matching objective for extracting the minor subspace, Minor Subspace  Similarity Matching (MSSM). Moreover, we derive an adaptive MSSM algorithm that naturally maps onto a novel neural network with local learning rules and gives numerical results showing that our method converges at a competitive rate.  
\end{abstract}
\begin{keywords}
artificial neural networks, minor subspace analysis, dimensionality reduction.
\end{keywords}
\section{Introduction}
\label{sec:intro}

One of the most straightforward tasks that a neural network (NN) can perform is to learn a low-dimensional space of stimulus features that captures the directions of the lowest variance, forming the minor subspace (Fig.~\ref{figur:Summary}B).  This task is known as minor subspace analysis, and some datasets are well characterized by the directions of least variation \cite{williams2002products,welling2004extreme,weiss2007makes}. 
Minor subspace analysis (MSA) has been used for tasks such as total least square regression \cite{gao2015convergence}, direction of arrival estimation \cite{kong2010self}, and others \cite{kong2011dual,nguyen2013unified}. MSA is also integral to problems more closely related to neuroscience, such as invariance learning \cite{schraudolph1992competitive}, and slow features analysis \cite{wiskott2002slow}.

While online MSA algorithms with associated NNs exist \cite{luo1997minor,cirrincione2002mca,schraudolph1992competitive}, they are often the result of straightforward adaptations of Oja's learning rule for Principal Subspace Analysis (PSA). However, these NNs inherit the non-locality of the learning rules characteristic of the Oja's NNs. 
Besides understanding and modeling brain functions, increased biological realism in artificial NN can be useful for handling large datasets or streaming tasks \cite{giovannucci2018efficient}, and the development of neuromorphic hardware \cite{pehlevan2019spiking}. In particular, biologically plausible NNs operate online, i.e., sample-by-sample, avoiding storage of large datasets in memory. A necessary yet not sufficient learning for learning rules in biological networks is that the learning rules be local because synapses only have access to information about the neurons they connect.

Our two main contributions are the following. Firstly, to overcome the non-local nature of the rules used in existing NNs, we propose a similarity matching objective function for MSA. Secondly, we show that such an objective is optimized by an online algorithm that maps onto a neural network with local learning rules (Fig.~\ref{figur:Summary}C). Numerical experiments show that, despite using local learning rules, our neural network  performs competitively with existing methods that do not respect such constraints. 
\section{Background and Problem Setting}\label{sec:background}

Given $T$ centered input data samples $(\x_t)_{t=1}^T\in\R^{n}$, which defines the input matrix by $\X = [\x_1,\ldots,\x_T]\in \R^{n \times T}$, the MSA problem aims at finding a set of $m$ orthonormal vectors, denoted $\W = [\w_1,\ldots,\w_m]^\top \in \R^{m\times n},$ such that the projections of $\x_t$ onto these vectors, denoted by $(\y_t:=\W \x_t) \in \R^{m}$ has minimum variance. We also define the output matrix by $\Y = [\y_1,\ldots,\y_T]\in \R^{m \times T}$. The empirical covariance matrix of $\X$, denoted by $\C_{x}= \frac1T \X \X^\top$, is assumed to be full rank. One formulation of the MSA problem is thus
\begin{eqnarray}\label{eq:basic_mca}
\min_{\W \in \R^{m\times n}, \W \W^\top =\I_m } \frac{1}{2} \tr [\W \C_x \W^\top] ~~ .
\end{eqnarray}
Suppose the eigen-decomposition of $\C_{x} = \V_x {\bf \Lambda}_x \V_x^\top $, where ${\bf \Lambda_x } = \text{diag} (\lambda^x_1,\ldots,\lambda^x_n)$, with $\lambda^x_1 \geq \ldots \geq \lambda^x_n > 0$ are the eigenvalues of $\C_{x}$. 
It is well known that the optimal solution of the problem \eqref{eq:basic_mca} is the projections of the input dataset $\X$ onto its minor subspace. The minor subspace is  spanned by the columns of $\V_x$ corresponding to the $m$ smallest eigenvalues of $\C_x$, denoted by $\V^{MS}_m = [\v^x_{n-m+1},\ldots, \v^x_{n}]$. 
Standard singular value decomposition and other offline methods exist \cite{allaire2008numerical} to extract the $m$ minor subspace.

\subsection{Existing Learning Rules are Non-local}

Various NNs exist for solving MSA in the online setting. It is natural to identify the inputs $\x_t \in \R^n$ with the activity of $n$ upstream neurons at time, $t$. In response, the NN outputs an activity vector, $\y_t\in \Real^m$, with $m$ the number of output neurons. For each time $t$, $\y_t$ is obtained by multiplying $\x_t$ by the corresponding synaptic weights, $\W$ \cite{oja1992principal}. 
Existing learning rules used to train NNs for MSA result from adaptations of Oja's original work for principal subspace analysis (PSA). Indeed, PSA is a variance maximization problem formulated as
\begin{eqnarray}\label{eq:basic_pca}
\max_{\W \in \R^{m\times n}, \W \W^\top =\I_m } \frac{1}{2} \tr [\W \C_x \W^\top] ~~ .
\end{eqnarray}
Oja first proposed \cite{oja1992principal} a stochastic gradient ascent algorithm for solving PSA \eqref{eq:basic_pca} leading to the popular Oja's rule.

Thus, it appears natural to implement a stochastic gradient descent, instead of ascent, of the same objective function to obtain an algorithm for MSA. Oja algorithm for MSA is then
\begin{eqnarray}\label{eq:Oja}
\Delta \W  \approx - \eta \left(  \y_t \x_t^\top - \y_t \y_t^\top \W \right) ~~ ,
\end{eqnarray} 
with $\eta>0$ the learning rate. 
However, besides the fact that such an update rule leads to diverging weights \cite{oja1992principal}, implementing it in a single-layer NN architecture requires non-local learning rules \cite{pehlevan2019neuroscience}.  Indeed, the last term in \eqref{eq:Oja}  implies that updating the weight of a synapse requires knowledge of output activities of all other neurons, which are not available to the synapse.

\begin{figure}[!t]
\includegraphics[width=0.48\textwidth]{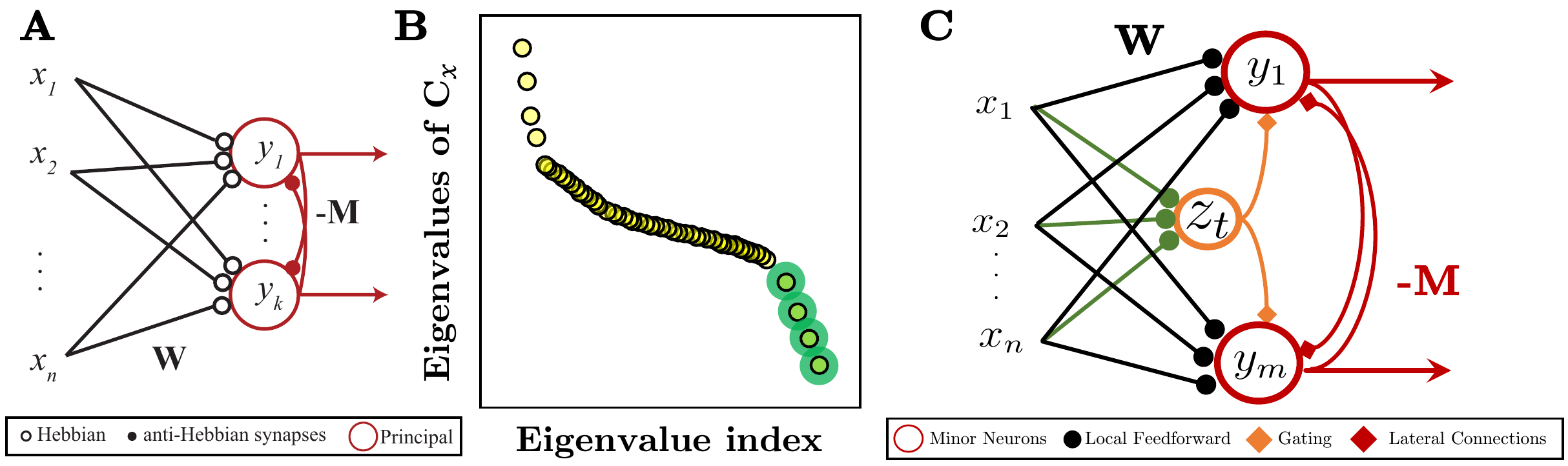} \caption{\textbf{(A)} Single-layer NN performing online PSA by  similarity matching \eqref{eq:sm-psp} \cite{pehlevan2015normative}. \textbf{(B)} Example of a sorted spectrum of an input covariance matrix, with four minor components highlighted. \textbf{(C)} Our proposed NN with local learning for MSA derived from MSSM \eqref{eq:sm_mca_costbis}. }
\vspace{-0.4cm}
\label{figur:Summary}
\end{figure}

\subsection{Similarity Matching for Principal Subspace Analysis}

To better understand our approach for building an MSA NN with local learning, we recall the similarity matching (SM) approach for deriving single-layer PSA NNs with local learning rules \cite{pehlevan2015normative}. If the similarity of a pair of vectors is quantified by their scalar product SM leads to the following objective:
\begin{eqnarray}
\label{eq:sm-psp}
\min_{\Y \in \R^{m\times T}} \frac{1}{T^2} \| \X^\top \X - \Y^\top \Y \|_F^2 ~~.
\end{eqnarray}
Despite a different form, both PSA \eqref{eq:basic_pca} and SM \eqref{eq:sm-psp} lead to the same embeddings \cite{cox2000multidimensional,williams2001connection}. Since $\C_x$ and $\X^\top \X$, have the same $n$ non-zero eigenvalues and related eigenvectors, SM also projects $\X$ onto the subspace of $m$ largest eigenvectors of $\C_x$.

A key insight of \cite{pehlevan2015normative,pehlevan2018similarity} was that the optimization problem \eqref{eq:sm-psp} can be converted algebraically to an online-tractable form by introducing dynamical variables $\W$ and $\M$: 
\begin{align}
\min_{\Y\in \Real^{m\times T}}\min_{\W\in \Real^{m\times n}}\max_{\M \in \Real^{m\times m}} \frac{1}{T} \Tr\left( -4\X^\top \W^\top \Y+ 2 \Y^\top\M \Y \right)\nonumber
\\ + 2 \Tr\left(\W^\top\W\right) - \Tr\left(\M^\top\M\right) .
\label{lyapunov}
\end{align}
They also proposed an online algorithm based on alternating optimization \cite{pehlevan2015normative} with respect to $\y_t$, and $(\W,\M)$ that can be implemented by a single-layer NN (Fig.~\ref{figur:Summary}A) as: 
\begin{align}
&\qquad \qquad \qquad \frac{d\y_t(\gamma)}{d\gamma}= \W \x_t - \M \y_t(\gamma) ~~,\label{grad}  \\
&\Delta \W  := \eta\left(\y_t \x_t^\top -\W \right) ~,~  \Delta \M := \eta\left(\y_t \y_t^\top -\M \right) . \label{Hebb}
\end{align}
As before, the activity of the upstream neurons encodes input variables, $\x_t$. Output variables, ${\bf y}_t$, are computed by the dynamics of activity \eqref{grad} in a single layer of neurons. They also suggested that the elements of matrices $\W$ and $\M$ are represented by the weights of synapses in feedforward and lateral connections, respectively. Crucially, unlike in \eqref{eq:Oja}, the resulting learning rules \eqref{Hebb} are local. 

\begin{figure}[!t]
\hspace{-0.65cm}\includegraphics[width=0.54\textwidth]{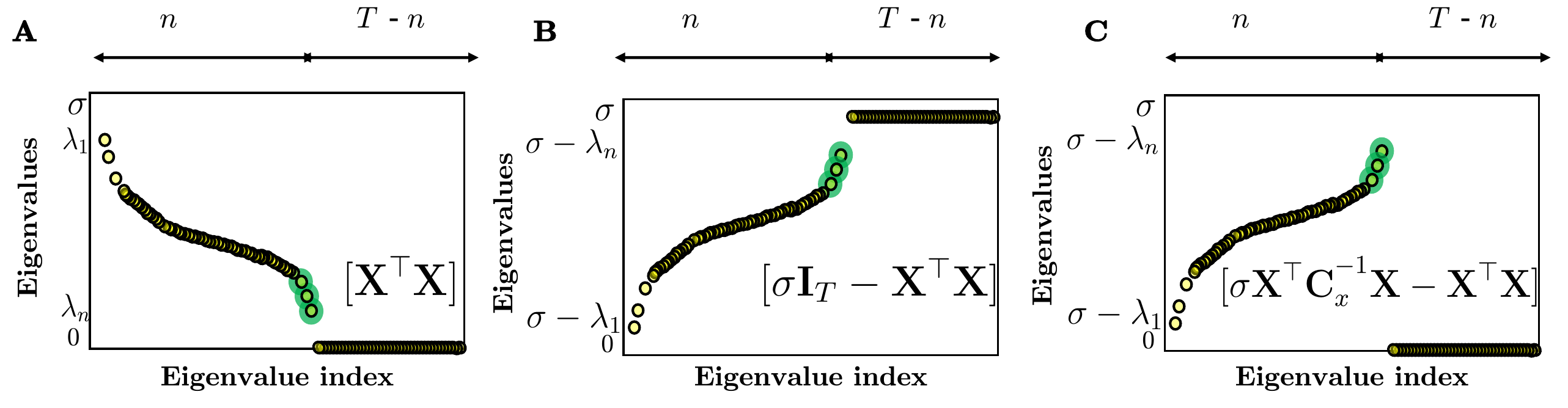}
\caption{Example of spectra of the different matrices of similarity considered. The eigenvalues are ordered according to the eigenvalue index of $\X^\top \X$. \textbf{(A)} shows the spectrum of $\X^\top\X$, and \textbf{(B)} of $[\sigma \I_T - \X^\top\X]$, and \textbf{(C)} $[\sigma \X^\top \C_x^{-1} \X - \X^\top \X]$ used in MSSM \eqref{eq:sm_mca_costbis}.} 
\label{figur:Spectrum_Gramians}
\vspace{-0.4cm}
\end{figure}

\section{A similarity matching approach to minor subspace analysis}
\label{sec:SM_MSA}

To overcome the non-locality of existing learning rules for MSA, we propose exploring a similarity matching approach. We present the first similarity matching objective function for MSA, referred to as Minor Subspace Similarity Matching (MSSM) in the following. We also derive an online algorithm for optimizing it.

\subsection{A Novel Objective Function for MSA} 

To develop our MSSM algorithm, we are looking for a similarity matrix with the following property: its eigenvectors associated with its largest eigenvalues must span the same subspace as the smallest non-zero eigenvalues of the original matrix of similarity $\X^\top\X$ (Fig.~\ref{figur:Spectrum_Gramians}A).

A similar problem was raised when considering the covariance matrix, $\C_x$, for which at least two methods exist for transforming the smallest eigenvalues into the largest eigenvalues. One is by considering the eigenvalue of $\C_x^{-1}$, the other is by considering $\sigma \I_n - \C_x$, with $\sigma>\lambda^x_1$. However, neither of these tricks work when the matrix of similarity $\X^\top \X$ is considered. 
Indeed, $\X^\top \X$ is not invertible if $T>n$, and has $T-n$ zero eigenvalues. Also, shifting the spectrum of $\X^\top \X$ by considering $\sigma \I_T - \X^\top \X$ makes the null eigenvalues of $\X^\top\X$ the largest eigenvalues of the resulting similarity matrix (Fig.~\ref{figur:Spectrum_Gramians}B), which is not the $m$ minor subspace of $\C_x$.

Let us now consider the matrix $\sigma \X^\top \C_x^{-1}\X$. The aforementioned matrix is the scaled matrix of similarity of whitened input. It has $n$ non-zero eigenvalues, all equal to $\sigma$, and the same eigenvectors as $\X^\top \X$. It is simply resulting from the fact that $\C_x^{-1/2}\X$ has all singular values equal to 1. 
Assuming that $\sigma>\lambda_1$, $\sigma \X^\top \C_{x}^{-1}\X  - \X^\top \X$, has the following spectrum, $\sigma -\lambda_n \geq \ldots \geq \sigma - \lambda_1 > 0 $, with $T-n$ null eigenvalues (Fig.~\ref{figur:Spectrum_Gramians}C). This matrix is thus the perfect candidate for the MSSM objective.

Our MSSM objective for discovering a low-dimensional subspace spanning the $m$-MS of $\C_x$, with $\sigma\geq \lambda_1$, is thus
\begin{align}\label{eq:sm_mca_costbis}
\min_{\Y\in\R^{m\times T}} \frac{1}{T^2} \norm{ \sigma \X^\top \C_x^{-1}\X - \X^\top \X - \Y^\top \Y}_F^2 ~~,  
\end{align}
from the following proposition with proof in Appendix~\ref{sec_SI:offline}. 
\begin{prop}\label{prop:sm_msp}
Optimal solutions $\Y^* \in \R^{m\times T}$ of MSSM \eqref{eq:sm_mca_costbis} are projections of the dataset $\X$ onto the $m$-dimensional minor subspace of $\C_x$, spanned by $\V^{MS}_x$ defined in Section~\ref{sec:background}
\end{prop}

\subsection{MSSM as a min-max optimization problem}

We propose a tractable min-max formulation of MSSM instrumental for deriving an algorithm that maps onto NN with local learning rules. 
We start by expanding the squared Frobenius norm and discard the terms independent of $\Y$ to obtain
\begin{align}\label{eq:sm_mca_cost}
    \min_{\Y\in\R^{m\times T}} -\frac{2}{T^2}\tr\left(\X^\top (\sigma \I_n - \C_x ) \C_x^{-1}\X \Y^\top\Y \right) \nonumber
    \\  \qquad \qquad \quad +\frac{1}{T^2}\tr\left(\Y^\top\Y\Y^\top\Y\right) ~~ .
\end{align}
We then introduce dynamical matrix variables $\W$, and $\M$ in place of $\sigma \frac1T\Y\X^\top \C_x^{-1}$, and $\frac1T\Y\Y^\top$, respectively. Similar substitution tricks are detailed in \cite{pehlevan2018similarity}. 
We can now rewrite \eqref{eq:sm_mca_cost} as the following min-max optimization problem 
\begin{align}\label{eq:offline_sm_mca_cost} 
\min_{\Y\in\R^{m\times T}}\min_{\W\in\R^{m\times n}}\max_{\M\in\mathbb{R}^{m\times m}} L(\W,\M,\Y)  
\end{align}
\begin{align*}
\vspace{-0.2cm}
\text{with}  ~~~~  L(\W,\M,\Y) :=  \frac1T \tr\left(-4 \X^\top (\sigma \I_n - \C_x) \W^\top \Y \right) \nonumber 
 \\ ~~ + \frac1T  \tr(  2 \Y^\top\M\Y) + \tr \left(2 \W \C_x \W^\top  - \M^\top \M\right). \nonumber
\end{align*}
In the offline setting, we can solve \eqref{eq:offline_sm_mca_cost} by alternating optimization \cite{olshausen1996emergence}. We first minimize with respect to $\Y$ while holding $(\W,\M)$ fixed, which admits a closed-form solution
\begin{align}
    \Y & =\M^{-1} \left( \sigma \W \X - \W \C_x \X \right)  ~~. 
\end{align}
Holding $\Y$ fixed, we then perform a gradient descent-ascent step with respect to $(\W, \M)$: 
\begin{align}
&\W \gets \W + 2\eta\left(\frac1T \Y ( \sigma \I_n - \C_x)  \X^\top  - \W \C_{x}\right) ~~  ; \\
&\M \gets \M+\frac\eta\tau\left(\frac1T \Y\Y^\top-\M\right) ~~  .
\end{align}
Here, $\eta>0$ is the learning rate for both $\W$, and $\tau>0$ is the ratio of the learning rates of $\W$ and $\M$. The stability of similar learning rules is investigated in \cite{pehlevan2015hebbian,lipshutz2020biologically}.

%
%
%
%

\subsection{Derivation of the Local Learning Rules}\label{subsec:online_MSA}

We now propose an online implementation of \eqref{eq:sm_mca_costbis} by observing that \eqref{eq:offline_sm_mca_cost} can be decomposed so that optimal outputs at different time steps can be computed independently as 
\begin{align}\label{eq:full_cost_online}
&\min_{\W \in\R^{m\times n}} \max_{\M\in\mathbb{R}^{m\times m}} \frac{1}{T} \sum_{t=1}^T  \left[ \tr \left( 2 \W \C_{x}\W^\top - \M^\top  \M \right) \right .  \nonumber
\\  
&\qquad \qquad \qquad \qquad \qquad \quad + \left. \min_{ {\bf y}_t\in \mathbb{R}^{m}}  l_t(\W,\M,\y_t) \right] 
\end{align}
\begin{align}\label{eq:neural_dyn_cost}
\text{with} ~~ l_t(\W,\M,\y_t)  := -4 z_t \x_t^\top \W^\top  \y_t + 2\y_t^\top \M \y_t ~~. 
\end{align}
with $z_t = ( \sigma  - \|\x_t\|^2)$. 
The approximation of $\C_x$ by $\x_t \x_t^\top$ is essential in the online setting as the true covariance matrix is not available and should be approximated at each $t$.

We can thus solve \eqref{eq:full_cost_online} sample-by-sample, i.e., online, by first minimizing \eqref{eq:neural_dyn_cost} with respect to the output variables, $\y_t$. To do so, we run the following neural dynamics obtained by gradient-descent until convergence, while keeping $(\W,\M)$ fixed: 
\begin{align}
\label{eq:neural_dyn}
    \frac{d\y_t(\gamma)}{d\gamma}= z_t \W \x_t -\M\y_t(\gamma) .
\end{align}
After the convergence of $\y_t$, we update $(\W,\M)$ by gradient descent-ascent as
\begin{eqnarray}
&\W &\gets ~ \W ~ + 2\eta_t \left(  z_t \y_t  -  \W \x_t\right) \x^\top_t \label{eq:local_hebb}   ~~, \\
&\M &\gets ~ \M ~~ + \frac{ \eta_t }{ \tau } (\y_t \y^\top_t -\M)    ~~  .  \label{eq:local_anti_hebb} 
\end{eqnarray}
Similarly to PSA SM, our algorithm can be implemented by a NN with feedforward, $\W$, and lateral connections $\M$ (Fig.\ref{figur:Summary}C). Here, however, the output and $\W$ update rules are gated by the global factor $z_t$. Global gating factors like $z_t$ have been used outside of the similarity matching framework for PCA and ICA \cite{isomura2018error,isomura2019multi}. 

\begin{figure*}[!t]
\hspace{-0.1cm}
\includegraphics[width=1.00\textwidth]{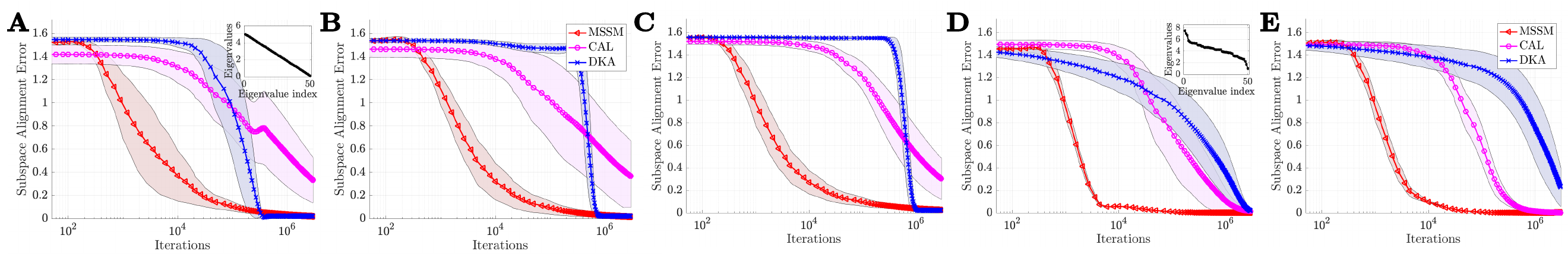}
\caption{Plot of the deviation of the MSSM, CAL, and DKA solutions from the $m$-MS, over 5 runs. 
Evaluation on the synthetic dataset with Linear spectrum for \textbf{(A)} $m=1$, \textbf{(B)} $m=2$, and \textbf{(C)} $m=4$. Evaluation on the synthetic dataset with Gaussian spectrum for \textbf{(D)} $m=1$, and \textbf{(E)}. Inset: Linear and Gaussian spectrum.} 
\label{figur:subspace_error}
\vspace{-0.3cm}
\end{figure*}
\section{Numerical Experiments}
\label{sec:numerics}

As an illustration of the capability of the NN derived from SMMS we provide experimental results of our algorithm against two popular algorithms proposed in \cite{chen1998unified} and \cite{douglas1998self}, denoted by \textbf{CAL} and \textbf{DKA}. The competing algorithms use the following update rules
\begin{align}
\label{eq:CAL_DKA}
\textbf{CAL} ~:~  \Delta \W  = - \eta  ( \W \W^\top \y\x^\top    -   \y\y^\top \W)  ~; \\
\textbf{DKA} ~:~ \Delta \W  = - \eta ( (\W \W^\top)^2 \y \x^\top    -   \y\y^\top \W ) ~ . 
\end{align}
We evaluate our algorithm on two artificially generated datasets, $\X \in \mathbb{R}^{50 \times 10,000}$, with a linear spectrum, ($\lambda_k=k/10, \forall k \in \{1,50\}$) (Fig.~\ref{figur:subspace_error}A,B and C), and with a randomly generated spectrum (Fig.~\ref{figur:subspace_error}D and E), respectively shown Inset of Fig.~\ref{figur:subspace_error}A and Fig.~\ref{figur:subspace_error}C.

The performance of the online algorithms are measured based on the subspace alignment error. Given matrices $(\W,\M)$, we define the projection $\F= \M^{-1}\W (\sigma \I_n - \C_x)$. The subspace alignment error is then generally defined by the relative difference in Frobenius norm square between the true normalized projector $\V^{MS}_m (\V^{MS\top}_m \V^{MS}_m)^{-1}\V^{MS\top}_m$ and the learned normalized projector $\F(\F^\top \F)^{-1}\F^\top$.

In Fig.~\ref{figur:subspace_error}, we show that after convergence, the subspace spanned by the synaptic connections learned by our online algorithms, is the same as the true basis vectors. In the experiments, our algorithm appears to be converging faster than CAL and DKA. 
However, no MSA algorithm, including those used here, have known provable convergence rates.

%
%

\section{Discussion}

In this work, we proposed a novel similarity matching objective function, and showed that the online optimization of such an objective leads to the extraction of the minor subspace of the input covariance matrix. The online algorithm we derived maps naturally onto a NN using only local learning rules.

Generalizing our work to the learning of other minor subspace analysis based tasks, such as slow feature analysis \cite{wiskott2002slow}, will open a path towards principled biologically plausible for invariance learning \cite{schraudolph1992competitive}, complementing the work on transformation learning from \cite{bahroun2019similarity}.

%

\appendix
\renewcommand{\theequation}{S.\arabic{equation}}
\setcounter{equation}{0}

\section{Proof of proposition }\label{sec_SI:offline}

Our result is an extension of the work of Mardia that connected PSA and similarity matching used in \cite{pehlevan2015hebbian} with proofs in \cite{mardia1980multivariate}. 
The result states that similarity matching is optimized by the projections of inputs onto the principal subspace of their covariance, i.e., performing PSA \cite{mardia1980multivariate,cox2000multidimensional}. 
%


\begin{propx}\label{prop:sm_psp}
For \X $\in \mathbb{R}^{n \times T}$, and fixed $m$ ($1\leq m \leq n)$, amongst all projections of $\X$ onto $m$-dimensionsional subspaces of $\mathbb{R}^n$, the objective \eqref{eq:sm-psp} is minimized when $\X$ is projected onto its principal coordinates in $m$ dimensions. (Mardia et al. \cite{mardia1980multivariate} Theorem 14.4.1). 
\end{propx}

%

Now, to show our results we need the following two results. Firstly, that $\X^\top \X$ and $\X^\top \C_x^{-1} \X$ are simultaneously diagonalizable in the basis formed by the eigenvector of $\X^\top \X$. Secondly, that the subspace associated with $m$ largest eigenvalues of $\sigma \X^\top \C_x^{-1} \X - \X^\top \X$ is the same as that the $m$-subspace by the $m$ smallest non-zero eigenvalues of $\X^\top \X$. 
Finally, we apply Prop.\ref{prop:sm_psp} to the MSSM similarity objective. 

\textbf{Step 1:} By the spectral theorem $\exists \U \in \mathcal{O}_T (\mathbb{R})$, composed of the eigenvectors of $\X^\top \X$, such that  $\X^\top \X = \U {\bf \Lambda} \U^\top $, with $ {\bf \Lambda}$ a diagonal matrix of eigenvalues of $\X^\top \X$ sorted by decreasing order. 
We can now show that $\X^\top \C_x^{-1} \X$ is diagonalizable in the basis formed by the columns of $\U$. Indeed, for all $i \in \{1,n\}$, $\u_i$ be the $i$-th column of $\U$ by definition we have that $\X^\top \X \u_i = \lambda_i \u_i$. As a result we have that
\begin{eqnarray}
\X^\top \C_x^{-1} \X \u_i &=& \frac{T}{\lambda_i} \X^\top \left[ \C_x^{-1} \frac{1}{T}\X \X^\top  \right] \X \u_i \nonumber \\
                          &=& \frac{T}{\lambda_i} \X^\top \X \u_i   = T \u_i ~~,
\end{eqnarray}
which proves that all eigenvectors of $\X^\top \X$ are eigenvectors of $\X^\top \C_x^{-1} \X$. We can now rewrite the difference between the two matrices of similarities in the basis of $\U$ as 
\begin{align}
\sigma \X^\top \C_x^{-1} \X - \X^\top \X =& \tilde{\U}
\begin{pmatrix}
\tilde{\sigma} - \lambda_n &  & \\ 
 & \ddots & & \\ 
 && \tilde{\sigma} - \lambda_1  && \\
 & & &0 & \\
 & & & & 0 
 \end{pmatrix} \tilde{\U}^\top \nonumber 
\end{align}
with $\tilde{\U}= [\u_n,\ldots,\u_1,\u_{n+1},\ldots,\u_T]$ and $\tilde{\sigma} =\sigma T$. 
We can then use Prop.\ref{prop:sm_psp} on the new similarity matrix  to prove Prop.\ref{prop:sm_msp}~.

\newpage

\bibliographystyle{IEEEbib}
\bibliography{strings}

\begin{thebibliography}{10}

\bibitem{williams2002products}
C.~K. Williams and F.~V. Agakov,
\newblock ``Products of {G}aussians and probabilistic minor component
  analysis,''
\newblock {\em Neural Computation}, vol. 14, no. 5, pp. 1169--1182, 2002.

\bibitem{welling2004extreme}
M. Welling, C. Williams, and F.~V. Agakov,
\newblock ``Extreme components analysis,''
\newblock in {\em Advances in Neural Information Processing Systems}, 2004, pp.
  137--144.

\bibitem{weiss2007makes}
Y. Weiss and W.~T. Freeman,
\newblock ``What makes a good model of natural images?,''
\newblock in {\em 2007 IEEE Conference on Computer Vision and Pattern
  Recognition}. IEEE, 2007, pp. 1--8.

\bibitem{gao2015convergence}
Y. Gao, X. Kong, C. Hu, H. Zhang, and L. Hou,
\newblock ``Convergence analysis of {M}{\"o}ller algorithm for estimating minor
  component,''
\newblock {\em Neural Processing Letters}, vol. 42, no. 2, pp. 355--368, 2015.

\bibitem{kong2010self}
X. Kong, C. Hu, and C. Han,
\newblock ``A self-stabilizing {MSA} algorithm in high-dimension data stream,''
\newblock {\em Neural networks}, vol. 23, no. 7, pp. 865--871, 2010.

\bibitem{kong2011dual}
X. Kong, C. Hu, and C. Han,
\newblock ``A dual purpose principal and minor subspace gradient flow,''
\newblock {\em IEEE Transactions on Signal Processing}, vol. 60, no. 1, pp.
  197--210, 2011.

\bibitem{nguyen2013unified}
T.~D. Nguyen and I. Yamada,
\newblock ``A unified convergence analysis of normalized {PAST} algorithms for
  estimating principal and minor components,''
\newblock {\em Signal processing}, vol. 93, no. 1, pp. 176--184, 2013.

\bibitem{schraudolph1992competitive}
N.~N. Schraudolph and T.~J. Sejnowski,
\newblock ``Competitive anti-{H}ebbian learning of invariants,''
\newblock in {\em Advances in Neural Information Processing Systems}, 1992, pp.
  1017--1024.

\bibitem{wiskott2002slow}
L. Wiskott and T.~J. Sejnowski,
\newblock ``Slow feature analysis: Unsupervised learning of invariances,''
\newblock {\em Neural computation}, vol. 14, no. 4, pp. 715--770, 2002.

\bibitem{luo1997minor}
F.-L. Luo, R. Unbehauen, and A. Cichocki,
\newblock ``A minor component analysis algorithm,''
\newblock {\em Neural Networks}, vol. 10, no. 2, pp. 291--297, 1997.

\bibitem{cirrincione2002mca}
G. Cirrincione, M. Cirrincione, J. H{\'e}rault, and S. Van~Huffel,
\newblock ``The {MCA EXIN} neuron for the minor component analysis,''
\newblock {\em IEEE Transactions on Neural Networks}, vol. 13, no. 1, pp.
  160--187, 2002.

\bibitem{giovannucci2018efficient}
A. Giovannucci, V. Minden, C. Pehlevan, and D.~B. Chklovskii,
\newblock ``Efficient principal subspace projection of streaming data through
  fast similarity matching,''
\newblock in {\em 2018 IEEE International Conference on Big Data (Big Data)}.
  IEEE, 2018, pp. 1015--1022.

\bibitem{pehlevan2019spiking}
C. Pehlevan,
\newblock ``A spiking neural network with local learning rules derived from
  nonnegative similarity matching,''
\newblock in {\em ICASSP 2019-2019 IEEE International Conference on Acoustics,
  Speech and Signal Processing (ICASSP)}. IEEE, 2019, pp. 7958--7962.

\bibitem{allaire2008numerical}
G. Allaire and S.~M. Kaber,
\newblock {\em Numerical linear algebra}, vol.~55,
\newblock Springer, 2008.

\bibitem{oja1992principal}
E. Oja,
\newblock ``Principal components, minor components, and linear neural
  networks,''
\newblock {\em Neural Networks}, vol. 5, no. 6, pp. 927--935, 1992.

\bibitem{pehlevan2019neuroscience}
C. Pehlevan and D.~B. Chklovskii,
\newblock ``Neuroscience-inspired online unsupervised learning algorithms:
  Artificial neural networks,''
\newblock {\em IEEE Signal Processing Magazine}, vol. 36, no. 6, pp. 88--96,
  2019.

\bibitem{pehlevan2015normative}
C. Pehlevan and D. Chklovskii,
\newblock ``A normative theory of adaptive dimensionality reduction in neural
  networks,''
\newblock in {\em Advances in Neural Information Processing Systems}, 2015, pp.
  2269--2277.

\bibitem{cox2000multidimensional}
T.~F. Cox and M.~A. Cox,
\newblock {\em Multidimensional scaling},
\newblock Chapman and hall/CRC, 2000.

\bibitem{williams2001connection}
C.~K. Williams,
\newblock ``On a connection between kernel {PCA} and metric multidimensional
  scaling,''
\newblock in {\em Advances in Neural Information Processing Systems}, 2001, pp.
  675--681.

\bibitem{pehlevan2018similarity}
C. Pehlevan, A.~M. Sengupta, and D.~B. Chklovskii,
\newblock ``Why do similarity matching objectives lead to
  {H}ebbian/anti-{H}ebbian networks?,''
\newblock {\em Neural Computation}, vol. 30, no. 1, pp. 84--124, 2018.

\bibitem{olshausen1996emergence}
B.~A. Olshausen and D.~J. Field,
\newblock ``Emergence of simple-cell receptive field properties by learning a
  sparse code for natural images,''
\newblock {\em Nature}, vol. 381, pp. 607--609, 1996.

\bibitem{pehlevan2015hebbian}
C. Pehlevan, T. Hu, and D.~B. Chklovskii,
\newblock ``A {H}ebbian/anti-{H}ebbian neural network for linear subspace
  learning: A derivation from multidimensional scaling of streaming data,''
\newblock {\em Neural computation}, vol. 27, no. 7, pp. 1461--1495, 2015.

\bibitem{lipshutz2020biologically}
D. Lipshutz, Y. Bahroun, S. Golkar, A.~M. Sengupta, and D.~B. Chkovskii,
\newblock ``A biologically plausible neural network for multi-channel canonical
  correlation analysis,''
\newblock {\em arXiv preprint arXiv:2010.00525}, 2020.

\bibitem{isomura2018error}
T. Isomura and T. Toyoizumi,
\newblock ``Error-gated hebbian rule: A local learning rule for principal and
  independent component analysis,''
\newblock {\em Scientific reports}, vol. 8, no. 1, pp. 1--11, 2018.

\bibitem{isomura2019multi}
T. Isomura and T. Toyoizumi,
\newblock ``Multi-context blind source separation by error-gated hebbian
  rule,''
\newblock {\em Scientific reports}, vol. 9, no. 1, pp. 1--13, 2019.

\bibitem{chen1998unified}
T. Chen, S.~I. Amari, and Q. Lin,
\newblock ``A unified algorithm for principal and minor components
  extraction,''
\newblock {\em Neural networks}, vol. 11, no. 3, pp. 385--390, 1998.

\bibitem{douglas1998self}
S.~C. Douglas, S.-Y. Kung, and S.-i. Amari,
\newblock ``A self-stabilized minor subspace rule,''
\newblock {\em IEEE Signal Processing Letters}, vol. 5, no. 12, pp. 328--330,
  1998.

\bibitem{bahroun2019similarity}
Y. Bahroun, A. Sengupta, and D.~B. Chklovskii,
\newblock ``A similarity-preserving network trained on transformed images
  recapitulates salient features of the fly motion detection circuit,''
\newblock in {\em Advances in Neural Information Processing Systems}, 2019, pp.
  14178--14189.

\bibitem{mardia1980multivariate}
K.~V. Mardia, J.~T. Kent, and J.~M. Bibby,
\newblock {\em Multivariate analysis (probability and mathematical
  statistics)},
\newblock Academic Press London, 1980.

\end{thebibliography}

\end{document}